\documentclass[runningheads]{llncs}
\usepackage{graphicx}
\usepackage{amsmath,amssymb} % define this before the line numbering.
\usepackage{color}

\usepackage{times}
\usepackage{xcolor}
\usepackage{subcaption}
\usepackage{tabu}

\begin{document}
\pagestyle{headings}
\mainmatter

%===========================================================
\title{End-to-end training of object class detectors\\for mean average precision}
\titlerunning{End-to-end training for mean average precision}
\authorrunning{P. Henderson, V. Ferrari}

\author{Paul Henderson \& Vittorio Ferrari}
\institute{University of Edinburgh}

\maketitle

\begin{abstract}
%The abstract should summarize the contents of the paper and should
%contain at least 70 and at most 300 words. It should be set in 9-point
%font size and should be inset 1.0~cm from the right and left margins.
We present a method for training CNN-based object class detectors directly using mean average precision (mAP) as the training loss, in a truly end-to-end fashion that includes non-maximum suppression (NMS) at training time.
This contrasts with the traditional approach of training a CNN for a window classification loss, then applying NMS only at test time, when mAP is used as the evaluation metric in place of classification accuracy.
However, mAP following NMS forms a piecewise-constant structured loss over thousands of windows, with gradients that do not convey useful information for gradient descent.
Hence, we define new, general gradient-like quantities for piecewise constant functions, which have wide applicability.
We describe how to calculate these efficiently for mAP following NMS, enabling to train a detector based on Fast R-CNN~\cite{girshick15iccv} directly for mAP.
This model achieves equivalent performance to the standard Fast R-CNN on the PASCAL VOC 2007 and 2012 datasets, while being conceptually more appealing as the very same model and loss are used at both training and test time.
\end{abstract}

\section{Introduction}

Object class detection is the task of localising all instances of a given set of object classes in an image.
Modern techniques for object detection~\cite{girshick15iccv,zhang15cvpr_bayes-struct-obj-detn,sermanet14iclr,girshick14cvpr} use a convolutional neural network (CNN) classifier~\cite{krizhevsky12nips,simonyan15iclr}, operating on object proposal windows~\cite{alexe10cvpr,uijlings13ijcv,zitnick14eccv}.
Given an image, they first generate a set of windows likely to include all objects, then apply a CNN classifier to each window independently.
The CNN is trained to output one score for each possible object class on each window, and an additional one for `background' or `no object'.
Such models are trained for window classification accuracy: the loss attempts to maximise the number of training windows for which the CNN gives the highest score to the correct class.
At test time, the CNN is applied to every window in a test image, followed by a non-maximum suppression processing stage (NMS). This eliminates windows that are not locally the highest-scored for a class, yielding the output set of detections.
%When applied to an image at test time, all windows are scored by the CNN, then for each object class independently, those windows that are not locally the highest-scored are dropped (\textit{i.e.\ } non-maximum suppression, NMS~\cite{felzenszwalb10pami}), to yield a final set of detections.
Typically, the performance of the detector is evaluated using mean average precision (mAP) over classes, which is based on the ranking of detection scores for each class~\cite{everingham10ijcv}.

Thus, the traditional approach is to train object detectors with one measure, classification accuracy over all windows, but test with another, mAP over locally highest-scoring windows.
While the training loss correlates somewhat with the test-time evaluation metric, they are not really the same, and furthermore, training ignores the effects of NMS. % except in the limit of a perfect detector...
%The model structure is also slightly different at training and test time, due to the added NMS stage when testing.
As such, the traditional approach is not true end-to-end training for the final \textit{detection} task, but for the surrogate task of window \textit{classification}.

In this work, we present a method for training object detectors directly using mAP computed after NMS as the loss.
This is in accordance with the machine learning dictum that the loss we minimise at training time should correspond as closely as possible to the evaluation metric used at test time.
It also fits with the recent trend towards training models end-to-end for their ultimate task, in vision~\cite{long15cvpr,vinyals15cvpr,pfister15iccv} and other areas~\cite{sutskever14nips,levine16jmlr}, rather than training individual components for engineered sub-tasks, and combining them by hand.

Directly optimising for mAP following NMS is very challenging for two main reasons:
(i) mAP depends on the global ordering of class scores for all windows across all images, and as such is piecewise constant with respect to the scores; and,
(ii) NMS has highly non-local effects within an image, as changing one window score can have a cascading effect on the retention of many other windows.
In short, we have a structured loss over many thousands of windows, that is non-convex, discontinuous, and piecewise constant with respect to its inputs.
Our main contribution is to overcome these difficulties by proposing new gradient-like quantities for piecewise constant functions, and showing how these can be computed efficiently for mAP following NMS.
This allows us to train a detector based on Fast R-CNN~\cite{girshick15iccv} in a truly end-to-end fashion using stochastic gradient descent, but with NMS included at training time, and mAP as the loss.

Experiments on the PASCAL VOC 2007 and 2012 detection datasets~\cite{everingham15ijcv} show that end-to-end training directly for mAP with NMS reaches equivalent performance to the traditional way of training for window classification accuracy and without NMS.
It achieves this while being conceptually simpler and more appealing from a machine learning perspective, as exactly the same model and loss are used at both training and test time.
Furthermore, our method is widely applicable on two levels: firstly, our loss is a simple drop-in layer that can be directly used in existing frameworks and models; secondly, our approach to defining gradient-like quantities of piecewise-constant functions is general and can be applied to other piecewise-constant losses and even internal layers. For example, using our method can enable training directly for other rank-based metrics used in information retrieval, such as discounted cumulative gain~\cite{jarvelin00sigir}.
Moreover, we do not require a potentially expensive max-oracle to find the most-violating inputs with respect to the model and loss, as required by \cite{yue07sigir,song16icml,zhang15cvpr_bayes-struct-obj-detn}.
%\vitto{and means that future layer-adders need only consider the effect of their modifications on detection performance, not interactions with the surrogate loss}
%\vitto{fewer engineering decisions}

\section{Background}

We recap here how NMS is performed (Sec.~\ref{sec:nms}) and mAP calculated (Sec.~\ref{sec:map}).
Then, we describe Fast R-CNN~\cite{girshick15iccv} in more detail (Sec.~\ref{sec:frcnn}), as it forms the basis for our proposed method.

\subsection{Non-maximum suppression (NMS)}\label{sec:nms}
Given a set of windows in an image, with scores for some object class, NMS removes those windows which are not locally the highest-scored, to yield a final set of detections~\cite{felzenszwalb10pami}.
%This avoids multiple detections of the same object instance. % VF: we do not really need a reason
Specifically, all the windows are marked as retained or suppressed by the following procedure:
first, the highest-scored window is marked as retained, and all those overlapping with it by more than some threshold (\textit{e.g.\ } 30\% in \cite{girshick15iccv,girshick14cvpr}) intersection-over-union (IoU) are marked as suppressed;
then, the highest-scored window neither retained nor suppressed is marked as retained, and again all others sufficiently-overlapping are marked as suppressed.
This process is repeated until all windows are marked as either retained or suppressed.
The retained windows then constitute the final set of detections.

\subsection{Mean Average Precision (mAP)}\label{sec:map}
\begin{figure}[t]
  \centering
  \includegraphics[width=0.7\textwidth]{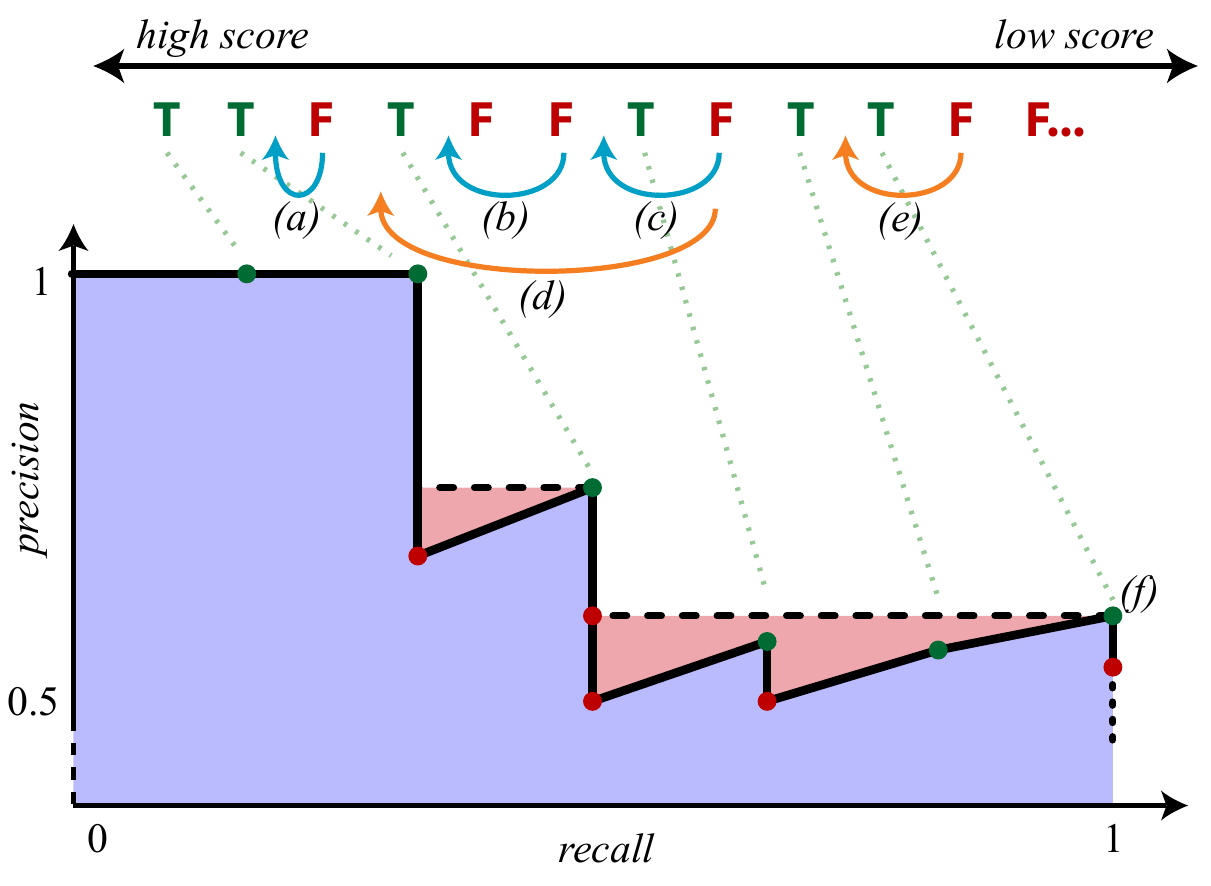}
  %\vskip -8pt
  \caption{Precision/recall curve (bottom) for a sequence of true-positive (TP) and false-positive (FP) detections ordered by score (top) for some object class with six ground-truth instances. Plotting the sequence of precision and recall values yields the black curve. The pink area shows the result of replacing each precision with the maximum at same or higher recall. AP is the total area of the pink and blue regions.
  The arrows (a-e) show the effect of positive perturbations to scores of FP detections. Blue arrows (a-c) show perturbations with no effect on AP: (a) the order of detections does not change; (b) the detection swaps places with another FP; (c) the detection swaps places with a TP, but a higher-recall TP (f) has higher precision so there is no change to area under the filled-in curve (pink shading). Orange arrows (d-e) show perturbations that do affect AP: (d) the same FP as (c) is moved beyond a TP that does appear on (hence affect) the filled in curve; (e) the FP moves past a single TP, altering the filled-in curve as far away as 0.5 recall.}
  \label{fig:curve-and-perturbations}
  %\vskip -16pt
\end{figure}

The mAP~\cite{everingham15ijcv,russakovsky15ijcv,everingham10ijcv} for a set of detections is the mean over classes, of the interpolated AP~\cite{salton86book} for each class.
This per-class AP is given by the area under the precision/recall (PR) curve for the detections (Fig.~\ref{fig:curve-and-perturbations}).

The PR curve is constructed by first mapping each detection to its most-overlapping ground-truth object instance, if any overlaps sufficiently---for PASCAL VOC, this is defined as overlapping with $>50\%$ IoU~\cite{everingham15ijcv}.
Then, the highest-scored detection mapped to each ground-truth instance is counted as a true-positive, and all other detections as false-positives.
Next, we compute recall and precision values for increasingly large subsets of detections, starting with the highest-scored detection and adding the remainder in decreasing order of their score.
Recall is defined as the ratio of true-positive detections to ground-truth instances, and precision as the ratio of true-positive detections to all detections.
The PR curve is then given by plotting these recall-and-precision pairs as progressively lower-scored detections are included.
Finally, dips in the curve are filled in (interpolated) by replacing each precision with the maximum of itself and all precisions occurring at higher recall levels (pink shading in Fig.~\ref{fig:curve-and-perturbations})~\cite{everingham10ijcv,salton86book}.
%\vitto{is this the exact protocol of the PASCAL VOC? Is this what is used also in ILSVRC-det? \pmh{yes, in all cases, but it is not stated in most of the papers!}}

The area under the interpolated PR curve is the AP value for the class.
For the PASCAL VOC 2007 dataset, this area is calculated by a rough quadrature approximation sampling at 11 uniformly spaced values of recall \cite{everingham10ijcv}; for the VOC 2012 dataset it is the true area under the curve \cite{everingham15ijcv}.

\subsection{Fast R-CNN}\label{sec:frcnn}

\begin{figure}[t]
  \centering
  %\vskip -13pt
  \begin{subfigure}{0.95\textwidth}
    \centering
    \includegraphics[width=0.75\textwidth]{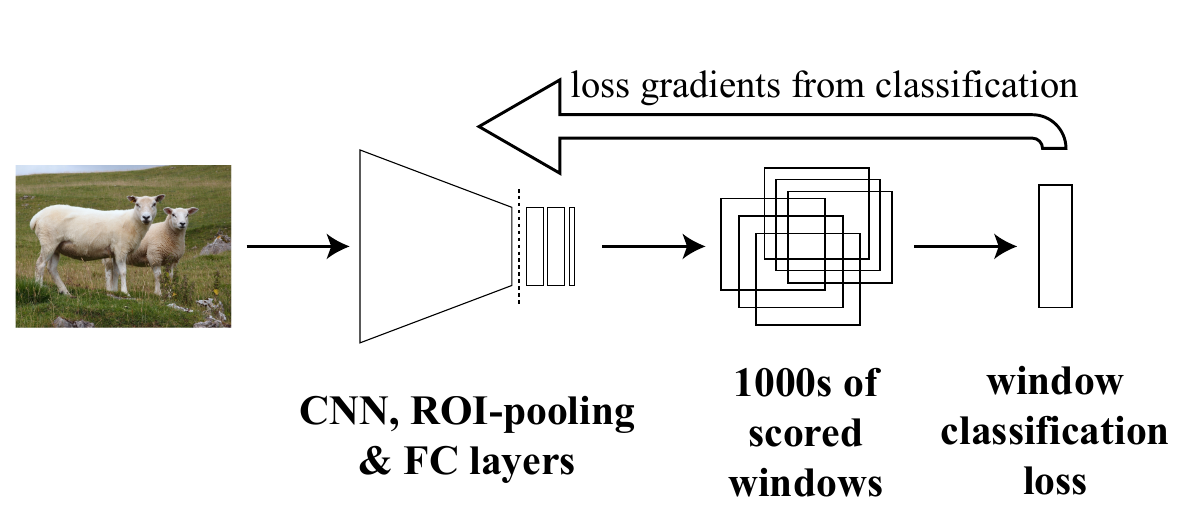}\vskip -8pt
    \caption{Fast R-CNN, training: scores passed directly to window classification loss}\label{fig:frcnn-train-arch}
  \end{subfigure}
  \vskip -12pt
  \begin{subfigure}{0.95\textwidth}
    \centering
    \includegraphics[width=1\textwidth]{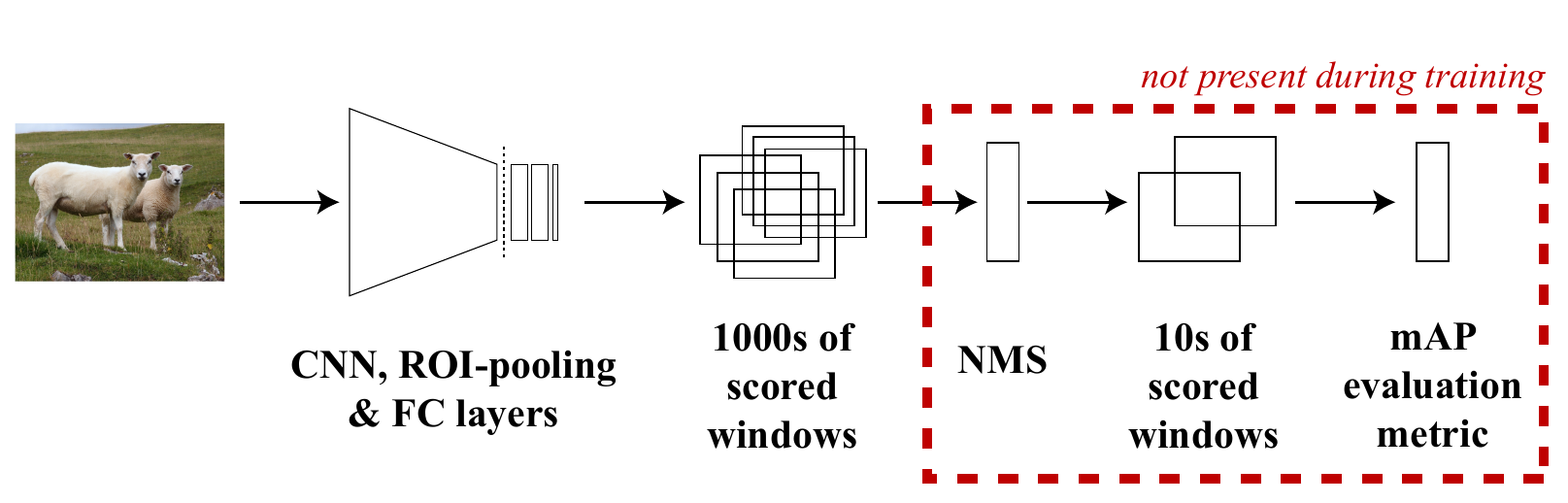}\vskip -8pt
    \caption{Fast R-CNN, testing: NMS applied, and detections evaluated with mAP}\label{fig:frcnn-test-arch}
  \end{subfigure}
  \begin{subfigure}{0.95\textwidth}
    \centering
    \includegraphics[width=0.95\textwidth]{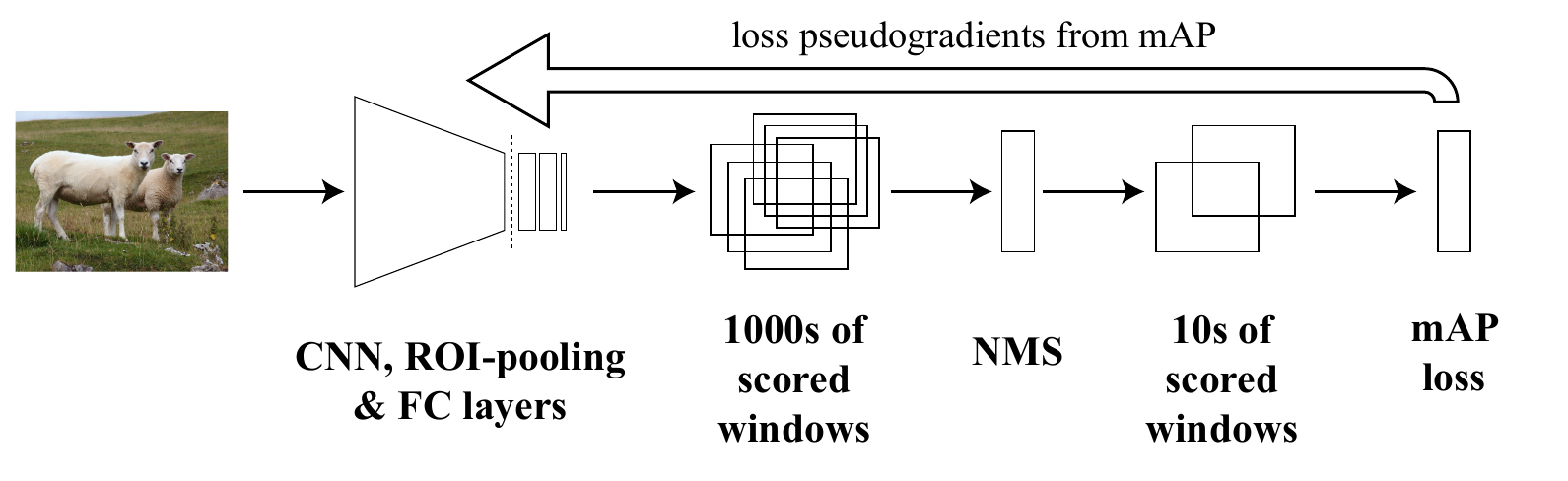}\vskip -8pt
    \caption{Our method, both training and testing: exactly the same operations occur at train and test time
    %\vitto{well, 'in the forward pass' (not backprop at test time)}
    , with identical model structure and the training loss matching the test-time evaluation metric}\label{fig:our-arch}
  \end{subfigure}
  \caption{Fast R-CNN~\cite{girshick15iccv} architecture during training (a) and testing (b) phases, and our architecture (c), which is the same in both phases.
  %In all cases, the image is passed through a CNN, features are extracted from regions, and scores are generated. % VF: not necessary, fig very big already
  }
  \label{fig:architectures}
  %\vskip -16pt
\end{figure}

\vskip 2pt\par\noindent\textbf{Model.}\hskip 0.5em
Our model is based on Fast R-CNN~\cite{girshick15iccv} (Figs.~\ref{fig:frcnn-train-arch}, \ref{fig:frcnn-test-arch}), without bounding-box regression.
This model operates by classifying proposal windows of an image, as belonging to one of a set of object classes, or as `background'.
Whole images are processed by a sequence of convolutional layers; then, for each window,
convolutional features with spatial support corresponding to that window are extracted and resampled to fixed dimension, before being passed through three fully-connected layers, the last of which yields a score for each object class and `background'.
The class scores for each window are then passed through a softmax function, to yield a distribution over classes.

\vskip 2pt\par\noindent\textbf{Training.}\hskip 0.5em
This network is trained with a window classification loss.
If a window overlaps a ground-truth object with IoU $> 0.5$, its true class is defined as being that object class; otherwise, its true class is `background'.
For each window, the network outputs softmax probabilities for each class, and the negative log likelihood (NLL) of the true class is used as the loss for that window; the total loss over a minibatch is simply a sum of the losses over all windows in it.
The network is trained by stochastic gradient descent (SGD) with momentum, operating on minibatches of two images at a time.

\vskip 2pt\par\noindent\textbf{Testing.}\hskip 0.5em
At test time, windows are scored by passing them forwards through the network, and recording the final softmax probabilities for each class.
Then, for each class and image, NMS is applied to the scored windows (Sec.~\ref{sec:nms}).
Note that this NMS stage is not present at training time.
Finally, the detections are evaluated using mAP over the full test set.

\section{Related Work}

%\vitto{generally the rel work section is rather long, at the end, when trying to fit, you could shorten most paragraphs}

Nearly all works on object class detection train a window classifier, and ignore NMS and mAP at training time.
Earlier approaches~\cite{felzenszwalb10pami,harzallah09iccv,dalal05cvpr,Viola01} apply the classifier to all windows in a dense regular grid, while more recently, object proposal methods~\cite{alexe10cvpr,zitnick14eccv} have been used to greatly reduce the number of windows~\cite{uijlings13ijcv,girshick14cvpr}.
Below we review the few works that try to either train for AP or other structured losses, or include NMS at training time.

%\pmh{
%other structured losses:
%-- mean iou for semantic seg, see holger
%-- mohapatra nips'14
%}

% parikh & zitnick nips wks 2011 notes that human-in-loop NMS is better than normal, indicating the is scope for improving NMS --- or as we do, at least accounting for it at training time

Blaschko et al.~\cite{blaschko08eccv} formulate object detection as a structured prediction problem, outputing a binary indicator for object presence and a set of bounding-box coordinates.
This is trained using a structured SVM, with a task loss that aims for correct classification and maximal IoU of predicted and ground-truth boxes in images containing the target class.
Like our method, this is a structured loss involving IoU of detections and ground-truth objects; however, it does not correspond to maximising AP, and only a single detection is returned in each image, so there is no NMS.
More recently, \cite{zhang15cvpr_bayes-struct-obj-detn} uses the same structured SVM loss, but with a CNN in place of a kernelised linear model over SURF features~\cite{blaschko08eccv}.
%Instead of a typical cutting-plane solver
This work directly optimises the structured SVM loss via gradient descent, allowing backpropagation to update the nonlinear CNN layers.

There exist works that train specifically for AP, but for classification problems, rather than for object detection with NMS. Yue et al.~\cite{yue07sigir} optimizes AP in the structured SVM framework---with a linear model, trained using a hinge loss weighted according to AP.
This requires solving a loss-augmented inference problem, \textit{i.e.\ } finding the scores that maximise the sum of AP and the output of the current model.
They present a dynamic programming algorithm to solve this, which has quadratic complexity in the number of training points.
Extending this work, \cite{song16icml} presents a more general technique for training nonlinear structured models directly for non-differentiable losses, again assuming that loss-augmented inference can be performed efficiently.
Using the same dynamic-programming approach as \cite{yue07sigir}, they apply it to the case of single-class AP with a model based on R-CNN~\cite{girshick14cvpr}, without NMS at training time.
While their method requires changes to the optimiser itself, ours does not.
Instead, we simply define a new loss layer that can be easily dropped into existing frameworks, and do not require solving a loss-augmented inference problem.
Furthermore, our approach can incorporate NMS and train simultaneously for multiple classes.
%and thus trains directly for mAP over object detections, as opposed to AP over binary window classification scores \pmh{clear?}.
Thus, while \cite{song16icml} trains for AP over binary window classification scores, ours trains directly for mAP over object detections. % \vitto{a bit repetitive, but ok}.

Taylor et al.~\cite{taylor08wsdm} discuss a different formulation for gradient-descent optimisation of certain losses based on ranking of scores (though not AP specifically).
They define a smooth proxy loss for a non-differentiable, piecewise constant ranking loss.
They treat the predicted score of each training point as a Gaussian random variable centered on the actual value, and hence compute the distribution of ranks for each score, by pairwise comparisons to all other scores.
This distribution is used in place of the usual hard ranks when evaluating the loss, and the resulting quantity is differentiable with respect to the original scores.
This method has cubic complexity in the number of training samples, making it intractable when there are tens of classes and thousands of windows (e.g. in PASCAL VOC).

Unlike most other approaches to object detection, \cite{wan15cvpr} includes NMS at training time as well as test time.
They use a deformable parts model over CNN features, that outputs scored windows derived from a continuous response map (in contrast to feeding fixed proposal windows through a CNN~\cite{girshick15iccv}).
The windows are passed through a non-standard variant of NMS.
Instead of training for mAP or window classification accuracy, the authors then introduce a new structured loss.
This includes terms for detections retained by NMS, but also for suppressed windows, in a fashion requiring knowledge of which detection suppressed them.
As such, it is deeply tied to the NMS implementation at training time, rather than being a generally-applicable loss such as mAP.

\section{Proposed Method}

%\begin{figure}[t]
%  \centering
%  \includegraphics[width=0.95\textwidth]{map-fd-train} \\
%  (a) training \\
%  \includegraphics[width=0.97\textwidth]{map-fd-test} \\
%  (b) testing
%  \caption{Architecture of our proposed method during training (a) and testing (b) phases. Note that the two models are identical, including the training loss matching the test-time evaluation metric, unlike Fast R-CNN (Fig.~\ref{fig:frcnn-arch}) \vitto{too big; shrink to single figure, and say that it's identical train/test}\pmh{merge as (c) of FRCNN figure?}}
%  \label{fig:our-arch}
%\end{figure}

We now describe our proposed method (Fig.~\ref{fig:our-arch}).
We discuss how our model differs from Fast R-CNN (Sec.~\ref{sec:our-model}) and why it is challenging to train (Sec.~\ref{sec:gradients}). Then we introduce our general method for defining gradients of piecewise-constant functions (Sec.~\ref{sec:pseudograd-theory}) and how we apply it to train our model (Sec.~\ref{sec:pseudograd-application}).

\subsection{Detection Framework}\label{sec:our-model}

\vskip 2pt\par\noindent\textbf{Model.}\hskip 0.5em
Our model is identical to Fast R-CNN as described above, up to the softmax layer: windows are still scored by passing through a sequence of convolutional and fully-connected layers.
As in \cite{girshick15iccv}, we can use different convolutional network architectures pretrained for ILSVRC 2012~\cite{russakovsky15ijcv} classification, such as AlexNet~\cite{krizhevsky12nips} or VGG16~\cite{simonyan15iclr}.
We omit the softmax layer, using the activations of the last fully-connected layer directly as window scores.
%Empirically, the softmax has little effect on the final performance \vitto{citation or corroborating evidence in our experiments} \pmh{could just omit mentioning performance non-impact} \vitto{ok, but you still wanna say it tends to saturate?} \pmh{well, it seems odd to leave it without justification}, but its tendency to saturate causes problems with propagating the loss gradients back through it.
% VF: as this point was not resolved by the Paul holiday point, I decided for some compromise version (below)
In our experiment we found that the softmax has little effect on the final performance, but its tendency to saturate causes problems with propagating the loss gradients back through it.
In contrast to Fast R-CNN, our model also includes an NMS layer immediately after the last fully-connected layer, which performs the same operation as used at test time for Fast R-CNN.
We regard the NMS layer as part of the model itself, present at both training and test time.

\vskip 2pt\par\noindent\textbf{Training.}\hskip 0.5em
During training, we add a loss layer that computes mAP over the minibatch, after NMS.
Thus, at training time, minibatches undergo exactly the same sequence of operations as at test time, and the training loss matches the test-time evaluation metric.
The network is still trained using SGD with momentum.
%however, the loss calculated by the NMS and mAP layers is non-convex and piecewise constant with respect to the class scores, so its gradients are everywhere zero or undefined.
%Hence, using the true gradient or subgradient does not yield useful results.
%In Sec.~\ref{sec:gradients}, we describe how to tackle this difficulty, while Sec.~\ref{sec:training-tricks} discusses some additional techniques used during training.
Section~\ref{sec:gradients} describes how to define derivatives of the mAP and NMS layers, while Sec.~\ref{sec:training-tricks} discusses some additional techniques used during training.

\vskip 2pt\par\noindent\textbf{Testing.}\hskip 0.5em
During testing, our method is identical to Fast R-CNN, except that the softmax layer is omitted.

\subsection{Gradients of mAP and NMS Layers}\label{sec:gradients}

In order to minimise our loss by gradient descent, we need to propagate derivatives back to the fully-convolutional layers of the CNN and beyond.
However, mAP is a piecewise constant function of the detection scores, as it depends only on their ordering---each score can be perturbed slightly without changing the loss.
The partial derivatives of such a loss function do not convey useful information for gradient descent (Fig.~\ref{fig:regular-gradient}) as they are almost everywhere zero (in the constant regions), and otherwise undefined (at the steps).
The subgradient is also undefined, as the function is non-convex.

Furthermore, even if we could compute the derivatives of mAP with respect to the class scores, they still need to be propagated back through the NMS layer.
This requires a definition of the Jacobian of NMS, which is again non-trivial.
Note that max-pooling layers are similarly non-differentiable, but good results are achieved by simply propagating the gradient back to the maximal input only.
We could do similar for NMS: allow only the locally-maximal windows propagate gradients back; however, this loses valuable information.
For example, if all detections overlapping some ground-truth object are suppressed, then there should be a gradient signal favouring increasing the score of those windows (or decreasing that of their suppressors). This does not occur if we na\"{i}vely copy gradients back through to maximal windows. %\pmh{and once again, there are more complex cases (like \textit{big} positive perturbations to suppressed windows), probably not worth mentioning}
In contrast, we require a Jacobian-like quantity for NMS that does capture this information.

We therefore develop general definitions for gradient-like quantities of piecewise-constant functions in Sec.~\ref{sec:pseudograd-theory}, and then describe how to apply them efficiently to NMS and mAP in Sec.~\ref{sec:pseudograd-application}.

\subsection{Pseudogradients of General Piecewise-Constant Functions}\label{sec:pseudograd-theory}

\begin{figure}[t]
  \centering
  \begin{subfigure}{0.45\textwidth}
    \centering
    \includegraphics[width=0.9\textwidth]{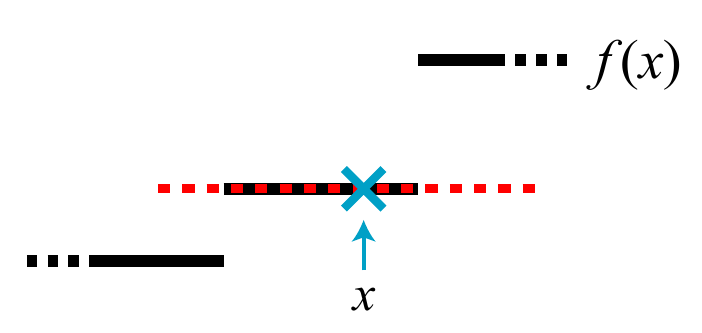}\vskip -8pt
    \caption{}\label{fig:regular-gradient}
  \end{subfigure}~~~
  \begin{subfigure}{0.45\textwidth}
    \centering
    \includegraphics[width=0.9\textwidth]{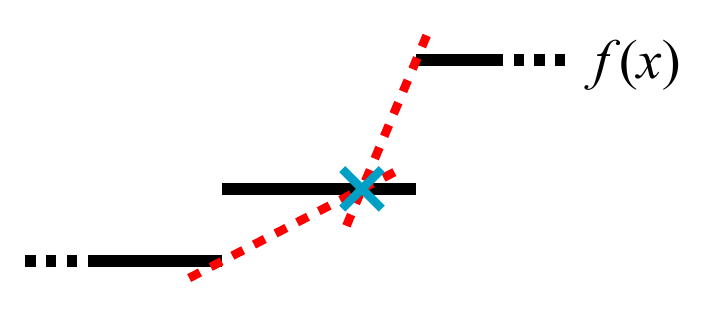}\vskip -8pt
    \caption{}\label{fig:sde-gradient}
  \end{subfigure} \\
  \begin{subfigure}{0.45\textwidth}
    \centering
    \includegraphics[width=0.9\textwidth]{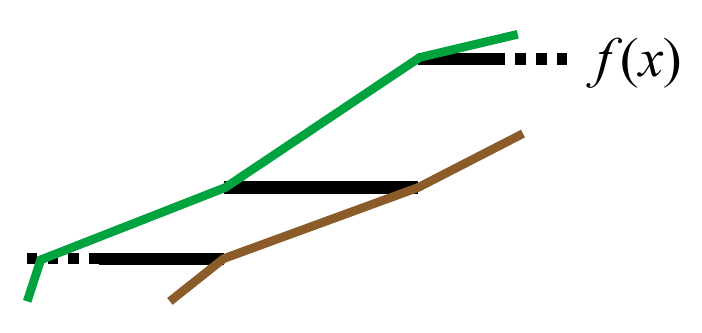}\vskip -8pt
    \caption{}\label{fig:envelopes}
  \end{subfigure}~~~
  \begin{subfigure}{0.45\textwidth}
    \centering
    \includegraphics[width=0.9\textwidth]{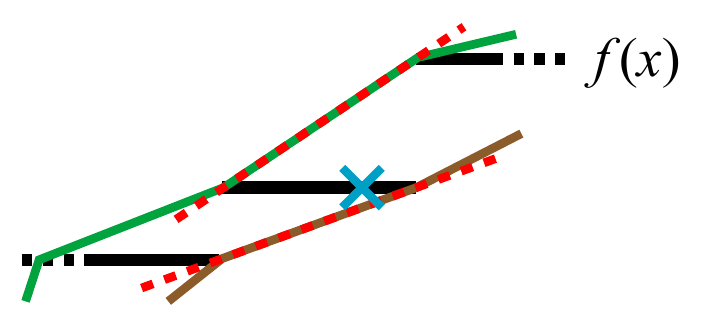}\vskip -8pt
    \caption{}\label{fig:lee-gradient}
  \end{subfigure}
  %\vskip -6pt
  \caption{A piecewise constant function $f(x)$ with steps at two points, and various definitions for gradients. (a) Conventional partial derivative (red dashed) at $x$, equal to zero, does not convey useful information for gradient descent. (b) Gradients at $x$ given by positive-perturbing and negative-perturbing finite difference estimators. (c) Piecewise-linear upper (green) and lower (brown) envelopes of $f(x)$. (d) Gradients at $x$ given by slope of upper/lower envelopes. When applied to our model, $f(x)$ is mAP, and the horizontal axis corresponds to the score of a single window with respect to which the partial derivative is being computed.} % PH-final: removed i.e.
  \label{fig:gradients}
  %\vskip -16pt
\end{figure}

We consider how to define a general \textit{pseudo partial derivative} (PPD) operation for piecewise-constant functions, that can be used to define quantities analogous to the gradient and the Jacobian.
For any piecewise-constant function $f(\mathbf{x})$ with countably many discontinuities (steps), we denote the PPD with respect to $x_i$ by $\widetilde{\partial}_{x_i} f$.
When the PPD is non-zero we need to move some non-infinitesimal distance before any change in the function occurs (unlike a conventional partial derivative). However when there is a change, it will be in the direction indicated by the PPD, and in magnitude corresponding to the PPD (this is made more precise below).
We then use our PPD to define an analogue to the gradient by $\widetilde{\nabla} f = (\widetilde{\partial}_{x_1} f, \ldots, \widetilde{\partial}_{x_N} f)$.
Intuitively, this tells us locally what direction to move so that the function will decrease, if we move some non-infinitesimal distance in this direction.
Similarly, for the Jacobian of vector-valued $\mathbf{f}$, we have $\widetilde{J}_{ij} = \widetilde{\partial}_{x_j} f_i$.

We now discuss two possible definitions for the PPD; these and the regular partial derivative are illustrated in Fig.~\ref{fig:gradients} for a one-dimensional function, at a point lying in a constant region between two steps.

\vskip 2pt\par\noindent\textbf{Finite difference estimators.}\hskip 0.5em
Most simply, we can apply a traditional single-sided finite difference estimator, as used for computing numerical gradients of a differentiable function.
Here, a small, fixed perturbation $\delta x$ is added to $x$, the function evaluated at this point, and the resulting slope used to approximate the gradient, by $\widetilde{\partial}_x f = \frac{f(x + \delta x) - f(x)}{\delta x}$.
The piecewise-constant functions we are interested in have finitely many steps, and so the probability of $f$ being undefined at the perturbed point is zero.
However, the constant regions of our function vary in size by several orders of magnitude, and so it is impossible to pre-select a suitable value for $\delta x$.
Instead, we use an adaptive approach: given $x$, set $\delta x$ to the smallest value such that $f(x + \delta x) \ne f(x)$, then compute $\widetilde{\partial}_x f$ as above (Fig.~\ref{fig:sde-gradient}).
Note that this method is single-sided: it only takes account of the change due to perturbing $x$ in one direction or the other.
This is undesirable, as in general, it delivers different results for each direction, perhaps yielding complementary information.
We address this issue by performing the same calculation independently with positive then negative perturbations $\delta x^+$ and $\delta x^-$, and taking a mean of the resulting pseudogradients.
We refer to this mean pseudogradient as SDE, for symmetric difference estimator.
This approach has the disadvantage that the magnitude of the gradient is sensitive to the exact location of $x$: if it is nearer to a step, the gradient will be larger, yet a correspondingly larger change to the network parameters may be undesirable.

\vskip 2pt\par\noindent\textbf{Linear envelope estimators.}\hskip 0.5em
An alternative approach to defining the PPD is to fit a piecewise-linear upper or lower envelope to the steps of the piecewise-constant function (Fig.~\ref{fig:envelopes}).
The PPD $\widetilde{\partial}_x f$ is then given by the slope of the envelope segment at the point $x$ (Fig.~\ref{fig:lee-gradient}).
In practice, we take the average of the gradients of the upper and lower envelopes.
Unlike SDE, this estimator does not become arbitrarily large as $x$ approaches a step.
%Note that this method also handles non-monotonic functions rather differently to SDE, as shown in \pmh{Fig.~\ref{fig:non-monotone}}. % in practice, our local approximation to NMS/mAP is monotonic wrt each coordinate, but that is irrelevant to this theoretical bit!
If $f$ has finitely many steps, then for all points before the first step and after the last, both linear envelopes have zero gradient; we find however that better results are achieved by using SDE in these regions, but with an empirically-tuned lower-bound on $\delta x$.
%\pmh{this kind of clipping is (implicitly) applied to fde too, as part of the general gdt-clipping we do, but philosophically that is a making-it-behave-numerically thing not a defining-the-gradient thing!}
We refer to this pseudogradient as MEE, for mean envelope estimator.

\subsection{Application to mAP and NMS}\label{sec:pseudograd-application}

To apply the above methods to mAP, we must compute the PPD of each class' AP with respect to each window score independently, holding the other scores constant.
This raises two questions:
(i) how to efficiently find the locations of the nearest step before and after a point, and
(ii) how to efficiently evaluate the loss around those locations.
We solve these problems by noting that changes to AP only occur when two scores change their relative ordering, and even then, only in certain cases.
Specifically, AP changes value only when a window counted as a true-positive changes place with one counted as a false-positive.
Also, the effective precision at a given recall is the maximum precision at that or any higher recall (Sec.~\ref{sec:map} and Fig.~\ref{fig:curve-and-perturbations}). So we have further conditions, 
\textit{e.g.\ } decreasing the score of a false-positive only affects AP when it drops below that of a true-positive at which precision is higher than any with even lower score.
This effect and other perturbations are illustrated in Fig.~\ref{fig:curve-and-perturbations} (blue and orange arrows).

\begin{figure}[t]
  \centering
  \begin{subfigure}{0.47\textwidth}
    \includegraphics[width=0.9\textwidth]{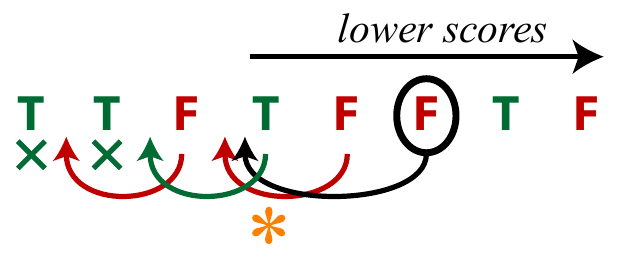}
    \vskip -8pt
    \caption{}
    \hfill
  \end{subfigure}
  \begin{subfigure}{0.47\textwidth}
    \vskip -7pt
    \hfill
    \includegraphics[width=0.9\textwidth]{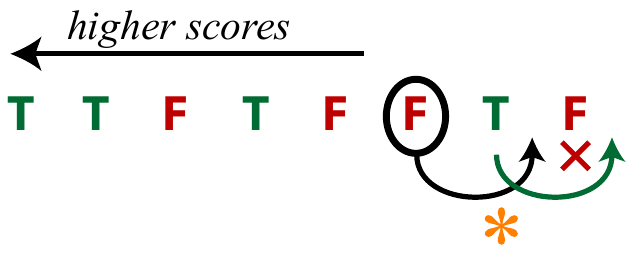}
    \vskip -8pt
    \caption{}
  \end{subfigure}
  %\vskip -10pt
  \caption{Efficient calculation of smallest perturbations to detection scores to cause a step in AP. In each case the circled FP is currently being considered. (a) Iterating detections in decreasing order of score, finding the smallest increase to each score that causes a change in AP (higher for TPs, lower for FPs). Detections already considered have an arrow showing where they are perturbed to; a cross indicates no increase to that score affects AP. When considering the circled FP, the last-seen TP is shown by the orange asterisk; perturbing the score of the circled detection just beyond (left) of this is the minimal change to affect AP. (b) Similar but iterating in increasing order of score, and hence calculating minimal decreases in score to affect AP.}
  \label{fig:perturbation-calculation}
  %\vskip -12pt
\end{figure}

Thus, for each class, we can find the nearest step before and after each point by making two linear passes over the detections, in descending then ascending order of score (Fig.~\ref{fig:perturbation-calculation}).
Assuming we have computed AP as described in Sec.~\ref{sec:map}, we know whether each detection is a true- or false-positive, and can keep track of the last-seen detection of each kind.
In the descending pass, for each detection, we find the smallest increase to its score that would result in a change to AP, thus giving the location of the nearest step on the positive side.
This score increase is that which moves it an infinitesimal amount higher than the score of the last-seen window of the other kind (true-positive vs. false-positive), subject to the additional conditions mentioned above.
Similarly, in the ascending pass, we can find the required decreases in scores that would cause a change in AP.
Once the step locations have been found, the new AP values resulting from perturbing the scores accordingly can be calculated by updating the relevant part of the PR curve, and then computing its area as normal.
Given the step locations and AP values, it is then straightforward to use the methods of Sec.~\ref{sec:pseudograd-theory} to compute the SDE or MEE.

\begin{figure}[t]
  \centering
  \begin{subfigure}{0.4\textwidth}
    \centering
    \includegraphics[width=0.6\textwidth]{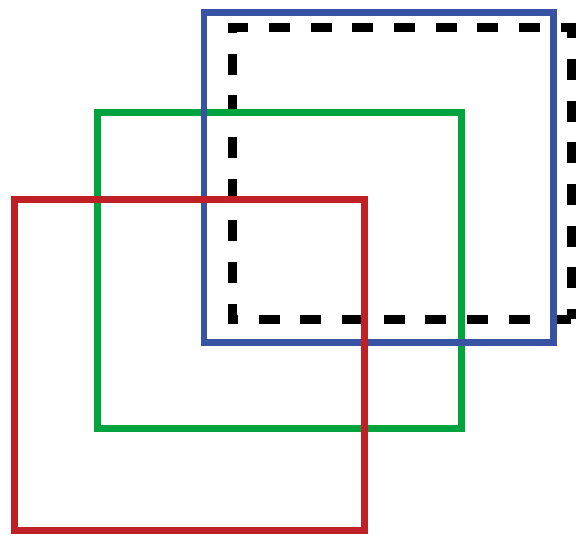}
    \vskip -4pt
    \caption{}
    \hfill
  \end{subfigure}
  \begin{subfigure}{0.4\textwidth}
    \centering
    \includegraphics[width=0.5\textwidth]{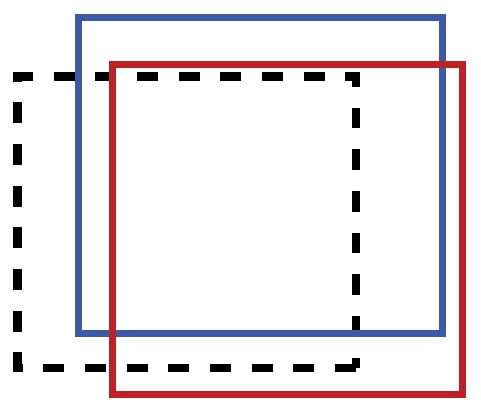}
    \caption{}
  \end{subfigure}
  %\vskip -10pt
  \caption{Transitivity approximations for NMS. Dashed black box is a ground-truth object; coloured boxes are scored windows, red $>$ green $>$ blue. (a) Red overlaps green sufficiently for NMS inhibition, and green overlaps blue similarly, but red does not overlap enough with blue. However, whether red is retained indirectly affects whether blue is retained, as if red suppresses green, then green does not suppress blue. In our approximation, this long-distance interaction between red and blue is ignored; however the two local interactions (red-green and green-blue) are included. (b) Red and blue overlap each other sufficiently for NMS inhibition; given that red suppresses blue, our approximation assumes that blue overlaps the same ground-truth instance as red (if any).}
  \label{fig:nms-approx}
  %\vskip -12pt
\end{figure}

\vskip 2pt\par\noindent\textbf{Incorporating NMS.}\hskip 0.5em
We must also account for NMS when propagating gradients back. The PPDs of NMS can be used to define a Jacobian as described in Sec.~\ref{sec:pseudograd-theory}, which may then be composed with the pseudogradient of mAP to define the gradient of mAP with respect to the pre-NMS scores.
%Most straightforwardly, derivatives of NMS can be computed independently from those of mAP.
%In that case, we can proceed similarly to max-pooling layers, so gradients propagate back only through locally-maximal windows.
%However, this is not sufficient for our purposes, as NMS may be suppressing all the windows overlapping some ground-truth instance, in which case there will be no gradient force to lower the score of the suppressing window or raise those of the suppressed ones.
However, subject to a small approximation, it is both easier and more efficient to consider NMS simultaneously with AP when determining step locations and the resultant changes to the loss.
Specifically, we introduce two transitivity approximations (Fig.~\ref{fig:nms-approx}):
(i) we do not attempt to model cascaded long-distance interactions between windows through multiple steps of NMS;
(ii) we assume in certain cases that windows suppressed by some detection overlap exactly the same ground-truth instances as the detection itself.
Under these approximations, it is possible to compute the PPDs with respect to pre-NMS scores in linear time in the number of windows.
This is achieved by:
(i) adding gradient contributions due to windows suppressed by a true-positive or false-positive detection at the same time as that detection, as these suppressed windows need to have their scores perturbed to the same point as their suppressor did to cause a change in AP;
(ii) including a third pass that adds gradient contributions from suppressed windows overlapping ground-truth instances that were missed entirely (\textit{i.e.\ } no detection covers them);
(iii) also adding gradient contributions from the detections that caused the suppressed-but-overlapping windows of (ii) to be suppressed.
%\vitto{soooo difficult! Maybe this paper would be a lot clearer with more math symbols, e.g. one to denote the set of all windows in an image, one for those that become detections, and one for those that are suppressed. Given those symbols, we could refer to these frequent objects compactly and clearly. Anyway, now it's too late for this submission. We might do it for the cam ready version (or resubm ;)}

\subsection{Training Protocol}\label{sec:training-tricks}

In order to train our model successfully, we make various changes to the training protocol used for Fast R-CNN in \cite{girshick15iccv}. The impact of each of these changes is given in Sec.~\ref{sec:experiments}.

\vskip 2pt\par\noindent\textbf{Minibatch composition.}\hskip 0.5em
We use larger minibatches than \cite{girshick15iccv}, as
(i) object detection mAP has a much higher batch-to-batch variance than simple window classification accuracy, and
(ii) including more windows increases the density of the gradient signal, as there are likely to be more false positives which score higher than some true positive (and vice versa).
%In particular, we draw windows \pmh{...}
%\vitto{how do we do the following? It does not come out by simply using larger minibatches! Also I am not convinced by the argument, I propose to cut this \pmh{I've clarified it, but still cut if you want}}
We also find that performance is improved by using proportionally fewer foreground windows (those overlapping a ground-truth instance as opposed to background) in each training minibatch.
While Fast R-CNN uses 25\% foreground windows, we use 5\%, which roughly corresponds to the distribution of windows seen at test time, when 5\% of all selective search proposals overlap a ground-truth instance.

\vskip 2pt\par\noindent\textbf{Regularisation.}\hskip 0.5em
Using our method, we found empirically that scores are prone to grow very large after several hundred iterations of training.
%\vitto{why? \pmh{my hypothesis (lacking direct evidence, hence I didn't include it): when softmax inputs (logits) get big for a TP, the softmax saturation prevents them from receiving any gradient signal to get even bigger, even if there are FPs with higher score and the softmax has not reached \textit{exactly} 1. Whereas in ours (mAP + no softmax), as long as there are higher-scored FPs somewhere in the minibatch, such a TP still gets a gradient signal to increase its score}}
This is effectively mitigated by introducing a regulariser on the window scores. We find that an L4 regulariser with very small weight performs best, as it gives greater freedom to smaller-magnitude scores while imposing a relatively hard constraint on magnitude, compared to the more common L1/L2 regularisation.

\vskip 2pt\par\noindent\textbf{Log-space.}\hskip 0.5em
We find it is beneficial to follow gradients of $\log(\text{mAP} + \epsilon)$ instead of mAP itself, for some small, fixed constant $\epsilon$.
Early in training when mAP is low, scores of true-positive windows are uniformly distributed amongst those of false-positive windows, and so an increase in the score of a true-positive often yields only a very small gain in mAP.
Using $\log(\text{mAP} + \epsilon)$ instead amplifies the effect of these changes, so training quickly escapes from the initial very low mAP.

\vskip 2pt\par\noindent\textbf{Gradient clipping.}\hskip 0.5em
%When using the adaptive finite difference method to compute gradients, the perturbation size $|\delta x|$ may be very small, and even otherwise, a small change in score may result in a disproportionately large step in AP.
%Hence, we find that numerical behaviour is improved (particularly at high learning rates) by employing gradient clipping.
We find that numerical behaviour is improved (particularly at high learning rates) by clipping elements of the gradient to a fixed threshold.

\section{Experiments}
\label{sec:experiments}

We now evaluate the performance of our approach on two datasets: PASCAL VOC 2007 and 2012~\cite{everingham15ijcv}.
Both datasets have 20 object classes; for VOC 2007, we train on the trainval subset (5011 images) and test on the test subset (4952 images); for VOC 2012, we train on the train subset (5717 images) and test on the validation subset (5823 images).
We also give results training on the union of VOC 2007 trainval and VOC 2012 trainval (total 16551 images), and testing on VOC 2007 test.
%\vitto{add a few stats, 20 classes, this many images, train, val, test subsets and how you use them \checkmark}

We compare our method to two others:
(i)
Fast R-CNN trained with the standard NLL loss for window classification, as described in \cite{girshick15iccv} (bounding box regression is disabled, to give a fair comparison with our method); and
(ii)
\cite{song16icml}, which also trains an R-CNN-like model for AP, but with a separate model for each class, no NMS at training time, and with a different way to compute parameter gradients. This is the closest work in spirit to ours.

\vskip 2pt\par\noindent\textbf{Settings.}\hskip 0.5em
We use Fast R-CNN as described in \cite{girshick15iccv}, built upon AlexNet~\cite{krizhevsky12nips} or VGG16~\cite{simonyan15iclr}, with weights initialised on ILSVRC 2012 classification~\cite{russakovsky15ijcv}.
We then remove the softmax layers at both training and test time, as described in Sec.~\ref{sec:our-model}, and replace the training loss layer with our NMS layer and mAP loss.

Incorporating the techniques described in Sec.~\ref{sec:training-tricks}, the overall loss we minimise by SGD is $L = -\log \left\{ \sum_c \text{AP}( \text{NMS}(\mathbf{s}_c)) / K \right\} + \lambda \sum_{c,b} |s_c^b|^4$, where $\mathbf{s}_c$ are
the window scores for class $c$, $K$ is the total number of classes, and $b$ indexes over windows.
%the pre-NMS scores for class $c$, $K$ is the total number of classes, and $b$ indexes over pre-NMS windows in a class. % PH: added for reviewer; not sure it actually makes things clearer! % PH-final: now precise-ified for VF

The AP calculation during training is always matched to that used for evaluation.
When testing on VOC 2007, we train using the VOC 2007 approximation to AP (Sec.~\ref{sec:map}); when testing on VOC 2012, we train using the true AP.
In order to compute pseudogradients for training, we try both SDE and MEE and compare their performance (Sec.~\ref{sec:pseudograd-theory}).
As our method works best with large minibatches, for the VGG16 experiments, we clamp the maximum image dimension to 600 pixels, to conserve GPU memory (this does not have a significant impact on the baseline performance).
%
%At test time, the only modification of Fast R-CNN is to use the pre-softmax scores as input to NMS. % VF: no need for this, confusing at this point, we said above we remove the softmax layers both at trn and test time

\begin{table}[t]
  \centering
  \caption{Performance of our method measured by mAP on VOC 2007 test set, with different pseudogradients (MEE vs SDE), network architectures (AlexNet vs VGG16), and training sets (VOC 2007 trainval vs union of VOC 2007 trainval and VOC 2012 trainval). We also give results for Fast R-CNN trained using a traditional softmax loss, without bounding box regression.}
  \label{tab:voc07}
  \vskip 4pt
  \renewcommand{\arraystretch}{1.15}
  \setlength{\tabcolsep}{4pt}
  \begin{tabu}{ccccc}
    \rowfont \itshape
    trained on... & \multicolumn{2}{c}{2007 only} & \multicolumn{2}{c}{2007 + 2012} \\
  			   & AlexNet & VGG16 & AlexNet & VGG16 \\
    \textbf{Ours, MEE}    & 51.6 & 58.9 & 54.9 & 62.5 \\
    \textbf{Ours, SDE}    & 51.3 & 60.7 & 54.8 & 62.3 \\
    \textbf{Fast R-CNN} & 52.0 & 62.4 & 53.8 & 63.5
  \end{tabu}
\end{table}

\vskip 2pt\par\noindent\textbf{Main results on VOC 2007.}\hskip 0.5em
Table \ref{tab:voc07} shows how our methods compare with Fast R-CNN, testing on the PASCAL VOC 2007 dataset.
Overall, our method achieves comparable performance to Fast R-CNN. %, and slightly higher than it when using AlexNet with a larger training set. % VF: no need to claim this; if a reviewer studies the table, it'll see pretty much everything is within 1%, most often with our method being a little worse than Fast R-CNN. If we claim this +1% now, it'll sound pushy
The results also show that using a larger training set (union of VOC 2007 and 2012 trainval subsets) increases performance by up to 3.6\% mAP, compared to training from VOC 2007 trainval alone. % VF: as noted by lots of previous works, but now with both Paul and Vitto on holiday let's skip citations ;)
This effect is significantly stronger for our method than for Fast R-CNN: for AlexNet, we gain 3.3\% mAP compared with 1.8\% for Fast R-CNN; for VGG16, we gain 3.6\% compared with 1.1\% for Fast R-CNN.
This indicates that our approach particularly benefits from more training data, possibly because optimising for mAP implies many comparisons between windows. % PH: note COCO doesn't support this currently
%\pmh{possibly because of the large minibatches}
%\pmh{possibly because we use few foreground boxes}
%\pmh{(I'm not quite convinced by any of those reasons...)}. % VF: I picked the one that sounded most closely related to our main contribution ;)
%
Of our two pseudo-gradient estimators, MEE slightly outperforms SDE, in all cases apart from VGG16 training on VOC 2007 trainval only.
This is likely because MEE is insensitive to the distances from points to nearest steps, in contrast to SDE (Sec.~\ref{sec:pseudograd-theory}); hence, MEE is a more robust
%\pmh{really I mean higher generalising power}
estimator of the impact of a score change, whereas SDE may introduce very large derivatives for a particular window.
%\pmh{unfortunately I have no idea why it does better than MEE on VGG VOC'07; it's likely an artefact due to interdependent/overfitted hyperparameters}
%\vitto{compare SDE and MEE along other axis, e.g. which one is faster? Simpler? Cooler-looking math? Memory consumption? Ask your girlfriend which acronym she likes more, ... \pmh{they are identical in terms of compute and memory, and very nearly so in terms of simplicity/maths...}}
%
In all cases, VGG16 significantly outperforms AlexNet, confirming previous studies~\cite{simonyan15iclr,girshick15iccv}.

\vskip 2pt\par\noindent\textbf{Ablation study.}\hskip 0.5em
In Sec.~\ref{sec:training-tricks}, we noted that certain modifications to the original training procedure of Fast R-CNN were necessary to achieve these results.
Ablating away these modifications reduces our mAP, as follows (all using AlexNet on VOC 2007 with the MEE gradient estimator):
%\begin{enumerate}
% \item 
(i)
   minibatch composition: increasing foreground fraction to 25\% (as used in Fast-RCNN): $-6.1$~mAP
% \item
(ii)
   minibatch size: halving batch size but doubling iteration count (so the same amount of data is seen): $-0.8$~mAP
% \item
(iii)
   score regularisation: with L2 regularisation instead of L4 and the constant adjusted appropriately: $-1.0$~mAP. With no regularisation, training fails after $<100$ iterations as the magnitude of the classification scores explode.
% \item
(iv)
   gradient clipping: with this disabled, training fails after $<100$ iterations due to numerical issues caused by large gradients.
%\end{enumerate}

\vskip 2pt\par\noindent\textbf{Comparison to \cite{song16icml} on VOC 2012.}\hskip 0.5em
The only previous work that attempts to train a CNN-based object detector directly for AP is \cite{song16icml}. % PH-final: clarified as CNN-based
Table \ref{tab:voc12-alexnet-vs-song} compares this method to ours; we use the PASCAL VOC 2012 dataset (testing on the validation subset) as this is what \cite{song16icml} reports results on.
Our method achieves comparable performance to \cite{song16icml}, with the MEE estimator again being slightly better than SDE.

Unlike our method, \cite{song16icml} trains a separate model for each class; their dynamic-programming solution to the loss-augmented inference problem is for single-class AP only (not mAP over all classes). Moreover, their training procedure does not take into account NMS.

\vskip 2pt\par\noindent\textbf{Discussion.}\hskip 0.5em
We hypothesise that our methods do not significantly outperform Fast R-CNN overall for three reasons.
%\begin{enumerate}
%  \item
(i)
  Our gradients are sparser than those of a softmax loss: not every window propagates information back for every class, as changing scores of certain windows has no effect on mAP (\textit{e.g.} low-scored background windows suppressed by NMS). For example, for VOC 2007, around 20\% of scores have a non-zero gradient --- compared with 100\% when using a softmax loss.
%  \item
(ii)
  mAP is a more rapidly changing function than the softmax loss: an estimate over a minibatch is a much higher-variance estimator of loss over the full set.
%  \item
(iii)
  It can be shown numerically that mAP over a minibatch of images is a biased estimator of mAP over the population of images from which that minibatch was drawn.
%\end{enumerate}

The real advantage of our method over the standard training procedure of Fast R-CNN is being more principled by respecting the theoretical need for having the same evaluation during training and test. 

\begin{table}[t]
  \centering
  \caption{Performance of our method compared with \cite{song16icml} (which trains for single-class AP, with a technique very different from ours). All models were trained on VOC 2012 train subset, tested on VOC 2012 validation subset, and use AlexNet. Bounding box regression was not used in any of the models.}
  \label{tab:voc12-alexnet-vs-song}
  \vskip 4pt
  \setlength{\tabcolsep}{8pt}
  \begin{tabu}{ccc}
    \rowfont \bfseries
    Ours, MEE & Ours, SDE & Song et al. \cite{song16icml} \\
    48.2 & 48.0 & 48.5
  \end{tabu}
\end{table}

%\pmh{perhaps analyse why we fail: although we're `good' in the sense of optimising the right thing, our gradient has high variance, and does not necessarily take all boxes into account effectively}

\section{Conclusions}

We have presented two definitions of pseudo partial derivatives of piecewise-constant functions.
Using these, we have trained a Fast R-CNN detector directly using mAP as the loss, with identical model structure at training and test time, including NMS during training.
This ensures that training is truly end-to-end for the final detection task, as opposed to window classification.
Our method achieves equivalent performance to Fast R-CNN.
It is easily integrated with standard frameworks for SGD, such as Caffe~\cite{jia13caffe}, as our NMS and mAP loss layers can be dropped in without affecting the minimisation algorithm or other elements of the model.
Our definitions of pseudogradients open up the possibility of training for other piecewise-constant losses.
In particular, ranking-based metrics are common in information retrieval, including simple AP on document scores, and discounted cumulative gain~\cite{jarvelin00sigir}.
Our method is very general as it does not require definition of an efficient max-oracle, in contrast to \cite{song16icml} and structured SVM methods.
Indeed, our approach can also be applied to piecewise-constant internal layers of a network, allowing back-propagation of gradients through such layers.

\bibliographystyle{splncs}
\bibliography{shortstrings,calvin,vggroup}

\end{document}